# RNN Approaches to Text Normalization: A Challenge

**Richard Sproat, Navdeep Jaitly**
Google, Inc.
{rws,ndjaitly}@google.com

## Abstract

This paper presents a challenge to the community: given a large corpus of *written* text aligned to its normalized *spoken* form, train an RNN to learn the correct normalization function. We present a data set of general text where the normalizations were generated using an existing text normalization component of a text-to-speech system. This data set will be released open-source in the near future.

We also present our own experiments with this data set with a variety of different RNN architectures. While some of the architectures do in fact produce very good results when measured in terms of overall accuracy, the errors that are produced are problematic, since they would convey completely the wrong message if such a system were deployed in a speech application. On the other hand, we show that a simple FST-based filter can mitigate those errors, and achieve a level of accuracy not achievable by the RNN alone.

Though our conclusions are largely negative on this point, we are actually not arguing that the text normalization problem is *intractable* using an pure RNN approach, merely that it is not going to be something that can be solved merely by having huge amounts of annotated text data and feeding that to a general RNN model. And when we open-source our data, we will be providing a novel data set for sequence-to-sequence modeling in the hopes that the the community can find better solutions.

## 1 Introduction

Within the last few years a major shift has taken place in speech and language technology: the field has been taken over by deep learning approaches. For example, at a recent NAACL conference well more than half the papers related in some way to word embeddings or deep or recurrent neural networks.

This change is surely justified by the impressive performance gains to be had by deep learning, something that has been demonstrated in a range of areas from image processing, handwriting recognition, acoustic modeling in automatic speech recognition (ASR), parametric speech synthesis for text-to-speech (TTS), machine translation, parsing, and *go* playing to name but a few.

While various approaches have been taken and some NN architectures have surely been carefully designed for the specific task, there is also a widespread feeling that with deep enough architectures, and enough data, one can simply feed the data to one's NN and have it learn the necessary function. For example:

> Not only do such networks require less human effort than traditional approaches, they generally deliver superior performance. This is particularly true when very large amounts of training data are available, as the benefits of holistic optimisation tend to outweigh those of prior knowledge. (Graves and Jaitly, 2014, page 1)

In this paper we present an example of an application that is unlikely to be amenable to such a "turn-the-crank" approach. The example is *text normaliza-*

*tion*, specifically in the sense of a system that converts from a written representation of a text into a representation of how that text is to be read aloud. The target applications are TTS and ASR — in the latter case mostly for generating language modeling data from raw written text. This problem, while often considered mundane, is in fact very important, and a major source of degradation of perceived quality in TTS systems in particular can be traced to problems with text normalization.

We start by describing why this application area is a bit different from most other areas of NLP. We then discuss prior work in this area, including related work on applications of RNNs in text normalization more broadly. We then describe a dataset that will be made available open-source as a challenge to the community, and we go on to describe several experiments that we have conducted with this dataset, with various NN architectures.

As we show below, some of the RNNs produce very good results when measured in terms of overall accuracy, but they produce errors that would make them risky to use in a real application, since in the errorful cases, the normalization would convey completely the wrong message. As we also demonstrate, these errors can be ameliorated with a simple FST-based filter used in tandem with the RNN.

But with a pure RNN approach, we have not thus far succeeded in avoiding the above-mentioned risky errors, and it is an open question whether such errors can be avoided by such a solution. We present below a hypothesis on why the RNNs tend to make the kinds of errors below. We close the paper by proposing a challenge to the community based on the data that we plan to release.

## 2 Why text normalization is different

To lay the groundwork for discussion let us consider a simple example such as the following:

```
A baby giraffe is 6ft tall and weighs
150lb.
```

If one were to ask a speaker of English to read this sentence, or if one were to feed it to an English TTS system one would expect that it to be read more or less as follows:

| a | <self> |
|---|--------|
| baby | <self> |
| giraffe | <self> |
| is | <self> |
| 6ft | six feet |
| tall | <self> |
| and | <self> |
| weighs | <self> |
| 150lb | one hundred fifty pounds |

Figure 1: Example input-output pairs for text normalization. In this example, the token `<self>` indicates that the token is to be left alone.

```
A baby giraffe is six feet tall and
weighs one hundred fifty pounds.
```

In the original written form there are two *non-standard words* (Sproat et al., 2001), namely the two measure expressions *6ft* and *150lb*. In order to read the text, each of these must be *normalized* into a sequence of ordinary words. In this case both examples are instances of the same *semiotic class* (Taylor, 2009), namely measure phrases. But in general texts may include non-standard word sequences from a variety of different semiotic classes, including measures, currency amounts, dates, times, telephone numbers, cardinal or ordinal numbers, fractions, among many others. Each of these involves a specific function mapping between the written input form and the spoken output form.

If one were to train a deep-learning system for text normalization, one might consider presenting the system with a large number of input-output pairs as in Figure 1. Here we use a special token `<self>` to indicate that the input is to be left alone. In principle this seems like a reasonable approach, but there are a number of issues that need to be considered.

The first is that one desirable application of such an approach, if it can be made to work, is to develop text normalization systems for languages where we do not already have an existing system. If one could do this, one could circumvent the often quite considerable hand labor required to build systems in more traditional approaches to text normalization.[1]

---

[1]E.g. Ebden and Sproat, 2014.

But where would one get the necessary data? In the case of machine translation, the existence of large amounts of parallel texts in multiple languages is motivated by the fact that people want to read texts in their own languages, and therefore someone, somewhere, will often go to the trouble of writing a translation. For speech recognition acoustic model training, one could in theory use closed captioning (Bharadwaj and Medapati, 2015), which again is produced for a reason. In contrast, there is no natural economic reason to produce normalized versions of written texts: no English speaker needs to be told that *6ft* is *six feet* or that *150lb* is *one hundred fifty pounds*, and therefore there is no motivation for anyone to produce such a translation. The situation in text normalization is therefore more akin to the situation with parsing, where one must create treebanks for a language in order to make progress on that language; if one wants to train text normalization systems using NN approaches, one must create the training data to do so. In the case of the present paper, we were able to produce a training corpus since we already had working text normalization systems, which allowed us to produce a normalized form for the raw input. The normalization is obviously errorful (we give an estimate of the percentage of errors below), but it is good enough to serve as a test bed for deep learning approaches — and by making it public we hope to encourage more serious attention to this problem. But in any event, if one is planning to implement an RNN-based approach to text normalization, one must take into consideration the resources needed to produce the necessary training data.

A second issue is that the set of interesting cases in text normalization is usually very sparse. Most tokens, we saw in the small example above, map to themselves, and while it is certainly important to get that right, one generally does not get any credit for doing so either. What is evaluated in text normalization systems is the interesting cases, the numbers, times, dates, measure expressions, currency amounts, and so forth, that require special treatment. Furthermore, the requirements on accuracy are rather stringent: if the text says *381 kg*, then an English TTS system had better say *three hundred eighty one kilograms*, or maybe *three hundred eighty one kilogram*, but certainly not *three hundred forty one kilograms*. *920* might be read as *nine hundred twenty*, or per-haps as *nine twenty*, but certainly never *nine hundred thirty*. *Oct 4* must be read as *October fourth* or maybe *October four*, but not *November fourth*. I mention these cases specifically, since the *silly* errors are errors that current neural models trained on these sorts of data will make, as we demonstrate below.

Again the situation for text normalization is different from that of, say, MT: in the case of MT, one usually must do something for any word or phrase in the source language. The closest equivalent of the `<self>` map in MT, is probably translating a word or phrase with its most common equivalent. This will of course often be correct: most of the time it would be reasonable to translate *the cat* as *le chat* in French, and this translation would count positively towards one's BLEU score. This is to say that in MT one gets credit for the "easy" cases as well as the more interesting cases — e.g. *this cat's in no hurry*, where a more appropriate translation of *cat* might be *type* or *gars*. In text normalization one gets credit only for the relatively sparse interesting cases.

Indeed, as we shall see below, if one is allowed to count the vast majority of cases where the right answer is to leave the input token alone, some of our RNNs already perform very well. The problem is that they tend to mess up with various semiotic classes in ways that would make them unusable for any real application, since one could never be quite sure for a new example that the system would not read it completely wrongly. As we will see below, the neural models occasionally read things like, *£900* as *nine hundred Euros* — something that state-of-the-art hand-built text normalization systems would never do, brittle though such systems may be. The occasional comparable error in an MT system would be bad, but it would not contribute much to a degradation of the system's BLEU score. Such a misreading in a TTS system would be something that people would immediately notice (or, worse, not notice if they could not see the text), and would stand out precisely because a TTS system ought to get such examples right.

In this paper we try two kinds of neural models on a text normalization problem. The first is a neural equivalent of a source-channel model that uses a sequence-to-sequence LSTM that has been successfully applied to the somewhat similar problem of grapheme-to-phoneme conversion (Rao et al., 2015), along with a standard LSTM language model architecture. The second treats the entire problem as a sequence-to-sequence task, using the same architecture that has been used for a speech-to-text conversion problem (Chan et al., 2016).

## 3 Prior work on text normalization

Text normalization has a long history in speech technology, dating back to the earliest work on full TTS synthesis (Allen et al., 1987). Sproat (1996) provided a unifying model for most text normalization problems in terms of weighted finite-state transducers (WFSTs). The first work to treat the problem of text normalization as essentially a language modeling problem was (Sproat et al., 2001). More recent machine learning work specifically addressed to TTS text normalization include (Sproat, 2010; Roark and Sproat, 2014; Sproat and Hall, 2014).

In the last few years there has been a lot of work that focuses on social media (Xia et al., 2006; Choudhury et al., 2007; Kobus et al., 2008; Beaufort et al., 2010; Kaufmann, 2010; Liu et al., 2011; Pennell and Liu, 2011; Aw and Lee, 2012; Liu et al., 2012a; Liu et al., 2012b; Hassan and Menezes, 2013; Yang and Eisenstein, 2013). This work tends to focus on different problems from those of TTS: on the one hand one, in social media one often has to deal with odd spellings of words such as `cu l8r`, `cooooooooooooollll`, or `dat suxxx`, which are less of an issue in most applications of TTS; on the other, expansion of digit sequences into words is critical for TTS text normalization, but of no interest to the normalization of social media texts.

Some previous work, also on social media normalization, that has made use of neural techniques includes (Chrupała, 2014; Min and Mott, 2015). The latter work, for example, achieved second place in the constrained track of the ACL 2015 W-NUT Normalization of Noisy Text (Baldwin et al., 2015), achieving an F1 score of 81.75%. In the work we report below on TTS normalization, we achieve accuracies that are comparable or better than that result (to the extent that it makes sense to compare across such quite different tasks), but we would argue that for the intended application, such results are still not good enough.

## 4 Dataset

Our data consists of 1.1 billion words of English text, and 290 million words of Russian text, from Wikipedia regions that could be decoded as UTF8, divided into sentences, and run through the Google TTS system's Kestrel text normalization system (Ebden and Sproat, 2014) to produce verbalizations. The format of the annotated data is as in Figure 1 above.

As described in (Ebden and Sproat, 2014), Kestrel's verbalizations are produced by first tokenizing the input and classifying the tokens, and then verbalizing each token according to its semiotic class. The majority of the rules are hand-built using the Thrax finite-state grammar development system (Roark et al., 2012). Statistical components of the system include morphosyntactic taggers for languages like Russian with complex morphology,[2] a statistical transliteration module (Jansche and Sproat, 2009), and a statistical model to determine if capitalized tokens should be read as words or letter sequences (Sproat and Hall, 2014). Most ordinary words are of course left alone (represented here as `<self>`), and punctuation symbols are mostly transduced to `sil` (for "silence").

The data were divided into 90 files (roughly 90%) for training, 5 files for online evaluation during training (the "development" set), and 5 for testing. In the test results reported below, we used the first 100K tokens of the final file (99) of the test data, including the end-of-sentence marker, working out to about 92K real tokens for English and 93K real tokens for Russian.

A manual analysis of about 1,000 examples from the test data suggests an overall error rate of approximately 0.1% for English and 2.1% for Russian. The largest category of errors for Russian involves years being read as cardinal numbers rather than the expected ordinal form.

Note that although the test data were of course taken from a different portion of the Wikipedia text than the training and development data, nonetheless a huge percentage of the individual tokens of the test

---

[2] The morphosyntactic tagger is an SVM model using hand-tuned features that classify the morphological bundle for each word independently, similar to SVMTool (Giménez and Màrquez, 2004) and MateTagger (Bohnet and Nivre, 2012).

data — 98.9% in the case of Russian and 99.5% in the case of English — were found in the training set. This in itself is perhaps not so surprising but it does raise the concern that the RNN models may in fact be *memorizing* their results, without doing much generalization. We discuss this issue further below.

Finally some justification of the choice of data is in order. We chose Wikipedia for two reasons. First, it is after all a reasonable application of TTS, and in fact it is used already in systems that give answers to voice queries on the Web. Second, the data are already publicly available, so there are no licensing issues.

## 5 Experiment 1: Text normalization using LSTMs

The first approach depends on the observation that text normalization can be broken down into two sub-problems. For any token:

- What are the possible normalizations of that token, and

- which one is appropriate to the given context?

The first of these — the **channel** — can be handled in a context-independent way by enumerating the set of possible normalizations: thus *123* might be *one hundred twenty three*, *one two three*, or *one twenty three*. The second requires context: in *123 King Ave.*, the correct reading in American English would normally be *one twenty three*.

The first component is a string-to-string transduction problem. Furthermore, since WFSTs can be used to handle most or all of the needed transductions (Sproat, 1996), the relation between the input and output strings is *regular*, so that complex network architectures involving, say, stacks should not be needed. For the input, the string must be in terms of *characters*, since for a string like *123*, one needs to see the individual digits in the sequence to know how to read it; similarly it helps to see the individual characters for a possibly OOV word such as *snarky* to classify it as a token to be left alone (`<self>`). On the other hand since the second component is effectively a **language-modeling** problem, the appropriate level of representation there is *words*. Therefore we also want the output of the first component to be in terms of words.

| John | `<self>` |
|------|----------|
| lives | `<self>` |
| at | `<self>` |
| 123 | one twenty three |
| King | `<self>` |
| Ave | avenue |
| near | `<self>` |
| A&P | a_letter and p_letter |
| . | sil |

Table 1: Training data format for the normalization channel, and language model. The channel model is trained to map from the first to the second column, whereas the language model is trained on the underlined tokens. The notation `x_letter` denotes a letter-by-letter reading, and `sil` denotes silence, which is predicted by the TTS text normalization system for most punctuation.

### 5.1 LSTM architecture

We train two LSTM models, one for the channel and one for the language model. The data usage of each during training is outlined in Table 1. For the channel model, the LSTM learns to map from a sequence of characters to one or more word tokens of output. For most input tokens this will involve deciding to leave it alone, that is to map it to `<self>`, or in the case of punctuation to map it to `sil`, corresponding to silence. For other tokens it must decide to verbalize it in a variety of different ways. For the language model, the system reads the words either from the input, if mapped to `<self>` or else from the output if mapped from anything else.

For the channel LSTM we used a bidirectional sequence-to-sequence model similar to that reported in (Rao et al., 2015) in two configurations: one with two forward and two backward hidden layers, henceforth the *shallow* model; and one with three forward and three backward hidden layers, henceforth the *deep* model. We kept the number of nodes in each hidden layer constant at 256.[3]. The output layer is a connectionist temporal classification (CTC) (Graves et al., 2006) layer with a softmax error function.[4] Input was limited to 250 distinct characters (including the unknown token). For the output, 1,000 distinct words (including `<self>` and the unknown token) were allowed for English, and 2,000 for Russian, the larger number for Russian being required to allow for various inflected forms. The number of words may seem small but it is sufficient to cover the distinct words that are output by the verbalizer for the various semiotic classes where the token does not simply map to `<self>`. See Figure 2.

---

[3] A larger shallow model with 1024 nodes in each layer ended up severely overfitting the training data.

[4] Earlier experiments with non-CTC architectures did not produce results as good as what we obtained with the CTC layer.

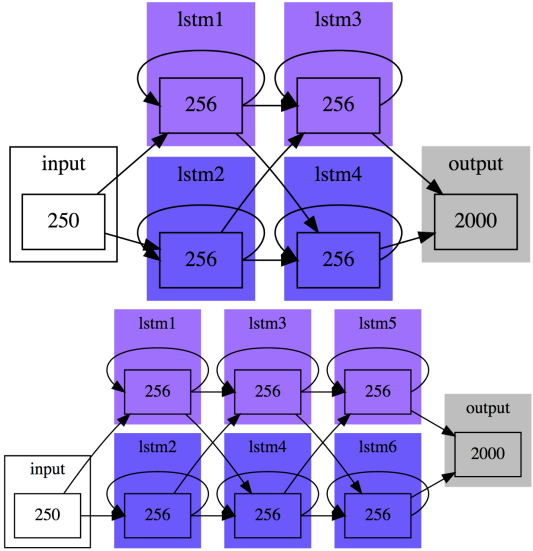

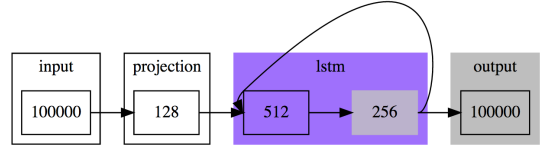

Figure 3: LSTM for the language model.

| | # Steps | Perp. | LER |
|---|---|---|---|
| Ru LM | 20.4B | 27.1 (*50.0*) | — |
| En LM | 17.4B | 42.7 (*64.8*) | — |
| Ru shal. chan. | 8.7B | — | 2.03% |
| Ru deep chan. | 6.5B | — | 1.97% |
| En shal. chan. | 8.3B | — | 1.75% |
| En deep chan. | 1.9B | — | 1.76% |

Table 2: Number of training steps, and perplexity or label error rate on held out data for the LSTMs. Note that the label error rate for the channel is calculated on the output word sequence produced by the model, including the `<self>` tag. For comparison the perplexities for a 5-gram WFST language model with Katz backoff trained using the toolkit reported in (Allauzen et al., 2016) on the same data and evaluated on the same held-out data are given in parentheses. Note that the language models were trained on 100 nodes for a total of 5 days and the channel models a total of 10 days.

Figure 2: LSTM architecture for the shallow and deep channel models for Russian. Purple LSTM layers perform forwards transitions and blue LSTM layers perform backwards transitions. The output is produced by a CTC layer with a softmax activation function. Input tokens are characters and output tokens are words. Numbers indicate the number of nodes in each layer.

The language model LSTM follows a standard RNN language model architecture, following on work of Mikolov et al., 2010, with an input and output layer consisting of $|V|$ nodes — we limited $|V|$ to 100,000, a dimensionality-reduction projection layer, a hidden LSTM layer with a feedback loop, and a hierarchical softmax output layer. During training the LSTM learns to predict the next word given the current word, but the feedback loop allows the model to build up a history of arbitrary length. See Figure 3.

The channel and language model LSTMs are trained separately. Table 2 shows the number of training steps and the final (dev) perplexity/label er-ror rate (LER) for the LM and channel LSTMs.

## 5.2 Decoding

At decoding time we need to combine the outputs of the channel and language model. This is done as follows. First, for each position in the output of the channel model, we prune the predicted output symbols. If one hypothesis has a very high probability (default 0.98), we eliminate all other predictions at that position: in practice this happens in most cases, since the channel model is typically very sure of itself at most of the input positions. We also prune all hypotheses with a low probability (default 0.05). Finally we keep only the $n$ best hypotheses *at each output position*: in our experiments we kept $n = 5$.

We then use the resulting pruned vectors to populate the corresponding positions in the input to the LM, with the channel probability for each token. This channel probability is multiplied by the LM probability times an LM weighting factor (1 in our experiments). This method of combining the chan-nel and LM probabilities can be thought of a as a poor-man's equivalent of the composition of a channel and LM weighted finite-state transducer (Mohri et al., 2002): the main difference is that there is no straightforward way to represent an arbitrary lattice in an LSTM, so that the representation is more akin to a confusion network ("sausage") at each output position.

## 5.3 Results

The first point to observe is that the overall performance is quite good: the accuracy is about 99% for English and 98% for Russian. But nearly all of this can be attributed to the model predicting `<self>` for most input tokens, and `sil` for punctuation tokens. To be sure these are decisions that a text normalization system must make: any such system must *decide* to leave an input token alone, or map it to silence. Still, as we noted in the introduction, one does not usually get credit for getting these decisions right, and when one looks at the interesting cases, the performance starts to break down, with the lowest performance predictably being found for cases such as TIME that are not very common in these data. The deep models also generally tend to be better than the shallow models, though this is not true in all cases — e.g. MONEY in English. This in itself is a useful sanity check, since we would expect improvement with deeper models.

The models are certainly able to get some quite complicated cases right. Thus for example for the Russian input *1 октября 1812 года* ("1 October, 1812"), the deep model correctly predicts *первого октября тысяча восемьсот двенадцатого года* ("first of october of the one thousand eight hundred and twelfth year"); or in English *2008-09-30* as *the thirtieth of september two thousand eight*.

But quite often the prediction is off, though in ways that are themselves indicative of a deeper problem. Thus consider the examples in Table 4, all of which are taken from the "deep" models for English and Russian. In both languages we find examples where the system gets numbers wrong, evidently because it is hard for the system to learn from the training data exactly how to map from digit sequences to number names — see also (Gorman and Sproat, 2016).[5] Other errors include reading the wrong unit as in the last three English examples, the reading of *hour* rather than *gigabyte* in the last Russian example, or the tendency of the Russian system to output `sil` in a lot of examples of measure phrases.

These errors are entirely due to the channel model: there is nothing ill-formed about the sequences produced, they just happen to be wrong given the input. One way to see this is to compute the "oracle" accuracy, the proportion of the time that the correct answer is in the pseudo-lattice produced by the channel. For the English deep model the oracle accuracy is 0.998. Since the overall accuracy of the model is 0.993, this means that $\frac{2}{7}$ or about 29% of the errors can be attributed to the channel model not giving the LM a choice of selecting the right model. What about the other 71% of the cases? While it is possible that some of these could be because the LM chooses an impossible sequence, in most cases what the channel model offers are perfectly possible verbalizations of *something*, just not necessarily correct for the given input. If the channel model offers both *twenty seconds* and *twenty kilograms*, how is the language model to determine which is correct, since even broader context may not be enough to determine which one is more likely?

The wrong-unit readings are particularly interesting in that one thing that is known to be true of RNNs is that they are very good at learning to cluster words into semantic groups based on contextual information. Clearly the system has learned the kinds of expressions that can occur after numbers, and some cases it substitutes one such expression for another. This property is the whole basis of word embeddings (Bengio et al., 2003) and other similar techniques. Interestingly, a recent paper (Arthur et al., 2016) discusses entirely analogous errors in neural machine translation, and attributes them to the same cause.[6] It is quite a useful property for many applications, but in the case of text normalization it is a drawback: unless one is trying to model a patient with semantic paraphasia, one generally wants a TTS system

---

[5] Note that in Russian most of the number errors involve the system reading the *correct number*, but in the wrong case form. These sorts of errors, while not desirable, are not nearly as bad as the system reading the wrong number: if all that is wrong is the inflection, a native speaker could still recover the intended meaning.

[6] Arthur et al.'s proposed solution is in some ways similar to the FST-based mechanism we propose below in Section 7.

| class | en shallow | | en deep | | ru shallow | | ru deep | |
|---|---|---|---|---|---|---|---|---|
| | N | cor | N | cor | N | cor | N | cor |
| ALL | 92448 | 0.990 | 92447 | 0.993 | 93194 | 0.982 | 93195 | 0.984 |
| PLAIN | 67954 | 0.997 | 67954 | 0.996 | 60702 | 0.998 | 60702 | 0.997 |
| PUNCT | 17729 | 1.000 | 17729 | 1.000 | 20264 | 1.000 | 20264 | 1.000 |
| DATE | 2808 | 0.858 | 2808 | 0.965 | 1495 | 0.886 | 1495 | 0.892 |
| TRANS | – | – | – | – | 4106 | 0.905 | 4106 | 0.919 |
| LETTERS | 1348 | 0.908 | 1348 | 0.939 | 1838 | 0.964 | 1838 | 0.971 |
| CARDINAL | 1069 | 0.936 | 1069 | 0.944 | 2388 | 0.771 | 2388 | 0.804 |
| VERBATIM | 1017 | 0.966 | 1017 | 0.969 | 1344 | 0.993 | 1344 | 0.993 |
| MEASURE | 142 | 0.908 | 142 | 0.944 | 411 | 0.623 | 411 | 0.645 |
| ORDINAL | 103 | 0.913 | 103 | 0.903 | 427 | 0.703 | 427 | 0.705 |
| DECIMAL | 89 | 0.989 | 89 | 0.978 | 60 | 0.617 | 60 | 0.583 |
| ELECTRONIC | 49 | 0.327 | 49 | 0.245 | 5 | 0.400 | 5 | 0.400 |
| DIGIT | 38 | 0.263 | 37 | 0.243 | 16 | 0.812 | 16 | 0.938 |
| MONEY | 37 | 0.946 | 37 | 0.892 | 18 | 0.500 | 19 | 0.474 |
| FRACTION | 16 | 0.562 | 16 | 0.562 | 23 | 0.565 | 23 | 0.478 |
| TIME | 8 | 0.625 | 8 | 0.750 | 8 | 0.625 | 8 | 0.875 |
| ADDRESS | 4 | 0.750 | 4 | 0.750 | – | – | – | – |

Table 3: Accuracies for the first experiment, including overall accuracy, and accuracies on various semiotic class categories of interest. Key for non-obvious cases: ALL = all cases; PLAIN = ordinary word (`<self>`); PUNCT = punctuation (`sil`); TRANS = transliteration; LETTERS = letter sequence; CARDINAL = cardinal number; VERBATIM = verbatim reading of character sequence; ORDINAL = ordinal number; DECIMAL = decimal fraction; ELECTRONIC = electronic address; DIGIT = digit sequence; MONEY = currency amount; FRACTION = non-decimal fraction; TIME = time expression; ADDRESS = street address. N is sometimes slightly different for each training condition since in a few cases the model produces no output, and we discount those cases — thus in effect giving the model the benefit of the doubt.

| Input | Correct | Prediction |
|---|---|---|
| 60 | sixty | six |
| 82.55 mm | eighty two point five five millimeters | eighty two one five five meters |
| 2 mA | two milliamperes | two units |
| £900 million | nine hundred million pounds | nine hundred million euros |
| 16 см | шестнадцати сантиметров | sil сантиметров |
| 16 cm | sixteen centimeters | sil centimeters |
| -11 | минус одиннадцать | минус один |
| | minus eleven | minus one |
| 100 000 | сто тысяч | один тысяч ноль ноль ноль ноль |
| | one hundred thousand | one thousand zero zero zero zero |
| 16 ГБ | шестнадцати гигабайтов | шестнадцать часов |
| | sixteen gigabytes | sixteen hours |

Table 4: Errors from the English and Russian deep models. In light of recent events, the final English example is rather amusing.

to read the written measure expression, not merely something from the same semantic category.

As we noted above, there was a substantial amount of overlap at the individual token level between the training and test data: could the LSTM simply have been memorizing? In the test data there were 475 unseen cases in English of which the system got 82.9% correct (compared to 99.5% among the seen cases); for Russian there were 1,089 unseen cases of which 83.8% were predicted correctly (compared to 98.5% among the seen cases). Some examples of the correct predictions are given in Table 5. As can be seen, these include some complicated cases, so it is fair to say that the system is *not* simply memorizing but does have some capability to generalize.

# 6 Experiment 2: Attention-based RNN sequence-to-sequence models

Our second approach involves modeling the problem entirely as a sequence-to-sequence problem. That is, rather than have a separate "channel" and language model phase, we model the whole task as one where we map a sequence of input characters to a sequence of output words. For this we use a Tensor Flow (Abadi et al., 2015) model with an attention mechanism (Mnih et al., 2014). Attention models are particularly good for sequence-to-sequence problems since they are able to continuously update the decoder with information about the state of the encoder and thus attend better to the relation between the input and output sequences. The Tensor Flow implementation used is essentially the same as that reported in (Chan et al., 2016).

In principle one might treat this problem in a way similar to how MT has been treated as a sequence-to-sequence problem (Cho et al., 2014), and simply pass the whole sentence to be normalized into a sequence of words. The main problem is that since we need to treat the input as a sequence of characters as we argued above, the input layer would need to be rather large in order to cover sentences of reasonable length. We therefore took a different approach and placed each token in a window of 3 words to the left and 3 to the right, marking the to-be-normalized token with a distinctive begin and end tag `<norm> ... </norm>`. Thus for example the token 123 in the context *I live at … King Ave .*

would appear as

```
I live at <norm> 123 </norm> King Ave .
```

on the input side, which would map to

```
one twenty three
```

on the output side.

In this way we were able to limit the number of input and output nodes to something reasonable. The architecture follows closely that of (Chan et al., 2016). Specifically, we used a 4-layer bidirectional LSTM reader (but without the pyramidal structure used Chan et al.'s task) that reads input characters, a layer of 256 attentional units, and a 2-layer decoder that produces word sequences. The reader is referred to (Chan et al., 2016) for more details of the framework.

It was noticed in early experiments with this configuration that the overabundance of `<self>` outputs was swamping the training and causing the system to predict `<self>` in too many cases. We therefore down-sampled the instances of `<self>` (and `sil`) in the training so that only roughly one in ten examples were given to the learner; among the various settings we tried, this seemed to give the best results both in terms of performance and reduced training time.

The training, development and testing data were the same as described in Section 4 above. The English RNN was trained for about five and a half days (460K steps) on 8 GPUs until the perplexity on the held-out data was 1.003; Russian was trained for five days (400K steps), reaching a perplexity of 1.002.

## 6.1 Results

As the results in Table 6 show, the performance is mostly better than the LSTM model described in Section 5. This suggests in turn that modeling the problem as a pure sequence-to-sequence transduction is indeed viable as an alternative to the source-channel approach we had taken previously.

Some errors are shown in Table 7. These errors are reminiscent of several of the errors of the LSTM system in Table 4, in that the wrong unit is picked. On the other hand it must be admitted that, in English, the only clear error of that type is the one example shown in Table 7. Again, as with the source-channel approach, there is evidence that while the system may be doing a lot of memorization, it is not merely memorizing. For English, 90.6% of the cases not found in the training data were correctly produced (compared to 99.8% of the seen cases); for Russian 86.7% of the unseen cases were correct (versus 99.4% of the seen cases). Complicated previously unseen cases in Russian, for example include examples like *9 июня 1966 г.*, correctly read as *девятое июня тысяча девятьсот шестьдесят шестого года* ('ninth of June, of the one thousand nine hundred sixty sixth year'); or *17.04.1750*, correctly read as *семнадцатое апреля тысяча семьсот пятидесятого года* ('seventeenth of April of the one thousand seven hundred fiftieth year').

| Input | Prediction |
|---|---|
| 13 October 1668 | the thirteenth of october sixteen sixty eight |
| 13.1549 km² | thirteen point one five four nine square kilometers |
| 26 июля 1864 | двадцать шестого июля тысяча восемьсот шестьдесят четвертого года |
| 26 July 1864 | twenty sixth of July of the one thousand eight hundred sixty fourth year |
| 90 кв. м. | девяносто квадратных метров |
| 90 sq. m. | ninety square meters |

Table 5: Correct output for test tokens that were never seen in the training data.

| | English | | Russian | |
|---|---|---|---|---|
| ALL | 92416 | 0.997 | 93184 | 0.993 |
| PLAIN | 68029 | 0.998 | 60747 | 0.999 |
| PUNCT | 17726 | 1.000 | 20263 | 1.000 |
| DATE | 2808 | 0.999 | 1495 | 0.976 |
| TRANS | – | – | 4103 | 0.921 |
| LETTERS | 1404 | 0.971 | 1839 | 0.991 |
| CARDINAL | 1067 | 0.989 | 2387 | 0.940 |
| VERBATIM | 894 | 0.980 | 1298 | 1.000 |
| MEASURE | 142 | 0.986 | 409 | 0.883 |
| ORDINAL | 103 | 0.971 | 427 | 0.956 |
| DECIMAL | 89 | 1.000 | 60 | 0.867 |
| ELECTRONIC | 21 | 1.000 | 2 | 1.000 |
| DIGIT | 37 | 0.865 | 16 | 1.000 |
| MONEY | 36 | 0.972 | 19 | 0.842 |
| FRACTION | 13 | 0.923 | 23 | 0.826 |
| TIME | 8 | 0.750 | 8 | 0.750 |
| ADDRESS | 3 | 1.000 | – | – |

Table 6: Accuracies for the attention-based sequence-to-sequence models. Classes are as in Table 3.

### 6.2 Results on "reasonable"-sized data sets.

The results reported in the previous sections depended on impractically large amounts of training data. To develop a system for a new language one needs a system that could be trained on data sets of a size that one could expect a team of native speakers to hand label. Assuming one is willing to invest a few weeks' of work with a small team, it is not out of the question that one could label about 10 million words of Wikipedia-style text.[7]

With this point in mind, we retrained the systems on 11.4 million tokens of English from the beginning of the original training set, and 11.9 million tokens of Russian. The system was trained for about 7 days for both languages, until the system had achieved a perplexity on held-out data of 1.002 and for Russian 1.007.

Results are presented in Table 8. The overall performance is not greatly different from the system trained on the larger dataset, and in some places is actually better.[8] The test data overlapped with the training data in 96.9% of the tokens for English and 95.5% for Russian, with the accuracy of the non-overlapped tokens being 95.0% for English and 93.5% for Russian.

The errors made by the system are comparable to errors we have already seen, though in English the errors in this case seem to be more concentrated in the reading of numeric dates. Thus to give just a few examples for English, reading *2008-07-28* as "the eighteenth of september seven thousand two", or *2009-*

| Input | Correct | Prediction |
|---|---|---|
| 2 mA | two milliamperes | two million liters |
| 11/10/2008 | the tenth of november two thousand eight | the tenth of october two thousand eight |
| 1/2 cc | half a c c | one minute c c |
| 18:00:00Z | eighteen hours zero minutes and zero seconds z | eighteen hundred cubic minutes |
| 55th | fifty fifth | five fifth |
| 750 вольт | семисот пятидесяти вольт | семьсот пятьдесят гектаров |
| 750 volts | seven hundred fifty volts | seven hundred fifty hectares |
| 70 градусами. | семьюдесятью градусами | семьюдесятью граммов |
| 70 degrees | seventy degrees | seventy grams |
| 16 ГБ | шестнадцати гигабайтов | шестнадцати герц |
| 16 GB | sixteen gigabytes | sixteen hertz |

Table 7: Errors from the attention sequence-to-sequence models.

| | English | | Russian | |
|---|---|---|---|---|
| ALL | 92416 | 0.996 | 93184 | 0.994 |
| PLAIN | 68029 | 0.997 | 60747 | 0.999 |
| PUNCT | 17726 | 1.000 | 20263 | 1.000 |
| DATE | 2808 | 0.974 | 1495 | 0.977 |
| TRANS | – | – | 4103 | 0.942 |
| LETTERS | 1404 | 0.974 | 1839 | 0.991 |
| CARDINAL | 1067 | 0.991 | 2387 | 0.954 |
| VERBATIM | 894 | 0.977 | 1298 | 1.000 |
| MEASURE | 142 | 0.958 | 409 | 0.927 |
| ORDINAL | 103 | 0.971 | 427 | 0.981 |
| DECIMAL | 89 | 1.000 | 60 | 0.917 |
| ELECTRONIC | 21 | 0.952 | 2 | 1.000 |
| DIGIT | 37 | 0.703 | 16 | 1.000 |
| MONEY | 36 | 0.972 | 19 | 0.895 |
| FRACTION | 13 | 0.846 | 23 | 0.739 |
| TIME | 8 | 0.625 | 8 | 0.750 |
| ADDRESS | 3 | 1.000 | – | – |

Table 8: Accuracies for the attention-based sequence-to-sequence models on smaller training sets. Classes are as in Table 3.

*10-02* as *the ninth of october twenty thousand two*. Some relatively complicated examples not seen in the training data that the English system got right included *221.049 km²* as *two hundred twenty one point o four nine square kilometers*, *24 March 1951* as *the twenty fourth of march nineteen fifty one* and *$42,100* as *forty two thousand one hundred dollars*.

Clearly then, the attention-based models are able to achieve with reasonable-sized data performances that are close to what it achieves with the large training data set. That said, the system of course continues to produce "silly" errors, which means that it will not be sufficient on its own as the text normalization component of a TTS system.

## 7 Finite-state filters

As we saw in the previous section, an approach that uses attention-based sequence-to-sequence models can produce extremely high accuracies, but is still prone to occasionally producing output that is completely misleading given the input. What if we apply some additional knowledge to filter the output so that it removes silly analyses?

One way to do this is to construct finite-state filters, and use them to guide the decoding. For example one can construct an FST that maps from expressions of the form `<number>` `<measure_abbreviation>` to a cardinal or decimal number and the possible verbalizations of the measure abbreviation. Thus *24.2kg* might verbalize as *twenty four point two kilogram* or *twenty four point two kilograms*. The

FST thus implements an overgenerating grammar that includes the correct verbalization, but allows other verbalizations as well.

We constructed a Thrax grammar (Roark et al., 2012) to cover MEASURE and MONEY expressions, two classes where the RNN is prone to produce silly readings. The grammar consists of about 150 lines, half of which consists purely mechanical language-independent rules to flip the order of currency expressions so that *£5* is transformed to its reading order *5£*. The Thrax grammar also incorporates English-specific lists of about 450 money and measure expressions so that, e.g., we know that *kg* can be *kilogram* or *kilograms*, as well as a number FST that is learned from a few hundred number names using the algorithm described in (Gorman and Sproat, 2016). Note that it is minimal effort to produce the lexical lists and the number name training data for a new language, certainly much less effort than producing a complete hand-built normalization grammar.[9]

During decoding, the FST is composed with the input token being considered by the RNN. If the composition fails — e.g. because this token is not one of the classes that the FST handles — then the decoding will proceed as it normally would via the RNN alone. If the composition succeeds, then the FST is projected to the output, and the resulting output lattice is used to restrict the possible outputs from the RNN. Since the input was a specific token — e.g. *2kg* — the output lattice will include only sequences that may be verbalizations of that token. This output lattice is transformed so that all prefixes of the output are also allowed (e.g. *two* is allowed as well as *two kilograms*). This can be done simply by walking the states in the lattice, and making all non-final states final with a free exit cost (i.e., 0 in the tropical semiring). However, we wish to give a strong *reward* for traversing the whole lattice from the initial to an original final state, and so the exit cost for original final states is set to a very low negative value (-1000 in the current implementation).[10]

The RNN decoder queries the lattice with a sequence of labels, the object being to find the possible transitions to the next labels and their cost. The label sequence is first transformed into a trivial acceptor, to which is concatenated an FSA that accepts any single output token, and thus has a branching factor of $|V|$, the size of the output vocabulary. This FSA is then composed with the lattice. For strings in the FSA that match against the lattice, the cost will be the exit cost at that state in the lattice; for strings that fail the cost will be `infinity`. Suppose that the input sequence is `two hundred`, and that the output lattice allows `two hundred kilogram` or `two hundred kilograms`. Then the FST will return a score of `infinity` for the label `milligram`, for example. However for `kilogram` or `kilograms` it will return a non-infinite cost, and indeed since exiting on one of these corresponds to an original final state of the grammar, it will accrue the reward discussed above. These costs will then be combined with the RNN's own scores for the sequence, and the final result computed as with the RNN alone. Note that since all prefixes of the sequences allowed by the grammar are also allowed, the RNN could, in the cited instance, produce *two hundred* as the output. However, it will get a substantial reward for finishing the sequence (*two hundred kilogram* or *two hundred kilograms*). As we shall see below, this is nearly always sufficient to persuade the RNN to take a more reasonable path.

We note in passing that this method is more or less the opposite approach to that of (Rastogi et al., 2016). In that work, the FST's scoring is augmented by an RNN, whereas in the present approach, the RNN's decoding is guided by the use of an FST.

Accuracies in English for the unfiltered and filtered RNN outputs, where the RNN is trained on the smaller training set described in the previous section, are given in Table 9. The MEASURE and MONEY sets show substantial improvement, while none of the other sets are affected, exactly as desired. Indeed, in this and the following tables we retain the scores for the non-MEASURE/MONEY cases in order to demonstrate that the performance on those classes is unaffected — not a given, since in principle the FSTs could overapply.

In order to focus in on the differences between the filtered and unfiltered models, we prepared a different subset of the final training file that was rich in MEASURE and MONEY expressions. Specifically, we selected 1,000 sentences, each of which had one expression in that category. We then decoded with the models trained on the smaller training set, with and without the FST filter. Results are presented in Table 10. Once again, the FST filter improves the overall accuracy for MONEY and MEASURE, leaving the other categories unaffected. Some examples of the improvements in both categories are shown in Table 11. Looking more particularly at measures, where the largest differences are found, we find that the only cases where the FST filter does not help is cases where the grammar fails to match against the input and the RNN alone is used to predict the output. These cases are *1/2 cc*, *30′* (for *thirty feet*), *80′*, *7000 hg* (which uses the unusual unit *hectogram*), *600 billion kWh* (the measure grammar did not allow for a spelled number like *billion*), and the numberless "measures" *per km*, */m²*. In a couple of other cases, the FST does not constrain the RNN enough: *1 g* still comes out as *one grams*, since the FST allows both it and the correct *one gram*, but this of course is an "acceptable" error since it is at least not misleading.

---

[9] The Thrax grammar and associated data will be released along with the main datasets.

[10] This final exit cost is actually set in the Thrax grammar itself, though it could as easily have been done dynamically at runtime.

| | RNN | | RNN+FST filter | |
|---|---|---|---|---|
| ALL | 92416 | 0.998 | 92435 | 0.998 |
| PLAIN | 68023 | 0.999 | 68038 | 0.999 |
| PUNCT | 17726 | 1.000 | 17729 | 1.000 |
| DATE | 2808 | 0.997 | 2808 | 0.997 |
| LETTERS | 1410 | 0.980 | 1411 | 0.980 |
| CARDINAL | 1067 | 0.995 | 1067 | 0.995 |
| VERBATIM | 894 | 0.985 | 894 | 0.985 |
| <span style="color:red">MEASURE</span> | <span style="color:red">142</span> | <span style="color:red">0.972</span> | <span style="color:red">142</span> | <span style="color:red">0.993</span> |
| ORDINAL | 103 | 0.990 | 103 | 0.990 |
| DECIMAL | 89 | 1.000 | 89 | 1.000 |
| ELECTRONIC | 21 | 1.000 | 21 | 1.000 |
| DIGIT | 37 | 0.838 | 37 | 0.838 |
| <span style="color:red">MONEY</span> | <span style="color:red">36</span> | <span style="color:red">0.972</span> | <span style="color:red">36</span> | <span style="color:red">1.000</span> |
| FRACTION | 13 | 0.846 | 13 | 0.846 |
| TIME | 8 | 0.750 | 8 | 0.750 |
| ADDRESS | 3 | 1.000 | 3 | 1.000 |

Table 9: Accuracies for the attention-based sequence-to-sequence models for English on smaller training sets, with and without an FST filter. (Slight differences in overall counts for what is the same dataset used for the two conditions, reflect the fact that a few examples are "dropped" by the way in which the decoder buffers data for the filterless condition.)

| | RNN | | RNN+FST filter | |
|---|---|---|---|---|
| ALL | 16160 | 0.997 | 16161 | 0.997 |
| PLAIN | 11224 | 0.999 | 11224 | 0.998 |
| PUNCT | 2585 | 1.000 | 2586 | 1.000 |
| DATE | 179 | 1.000 | 179 | 1.000 |
| LETTERS | 149 | 0.993 | 149 | 0.993 |
| CARDINAL | 434 | 0.998 | 434 | 0.998 |
| VERBATIM | 120 | 0.967 | 120 | 0.967 |
| <span style="color:red">MEASURE</span> | <span style="color:red">979</span> | <span style="color:red">0.979</span> | <span style="color:red">979</span> | <span style="color:red">0.991</span> |
| ORDINAL | 18 | 1.000 | 18 | 1.000 |
| DECIMAL | 132 | 1.000 | 132 | 1.000 |
| ELECTRONIC | 1 | 1.000 | 1 | 1.000 |
| DIGIT | 14 | 0.929 | 14 | 0.929 |
| <span style="color:red">MONEY</span> | <span style="color:red">312</span> | <span style="color:red">0.965</span> | <span style="color:red">312</span> | <span style="color:red">0.971</span> |
| FRACTION | 7 | 1.000 | 7 | 1.000 |
| TIME | 1 | 1.000 | 1 | 1.000 |
| ADDRESS | 2 | 1.000 | 2 | 1.000 |

Table 10: Accuracies for the attention-based sequence-to-sequence models for English on smaller training sets, with and without an FST filter, on the MEASURE-MONEY rich dataset.

Finally Table 12 shows results for the RNN with and without the FST filter on 1000 MONEY and MEASURE expressions that have not previously been seen in the training data.[11] In this case there was no improvement for MONEY, but there was a substantial improvement for MEASURE. In most cases, the MONEY examples that failed to be improved with the FST filter were cases where the filter simply did not match the input, and thus was not used.[12]

The results of a similar experiment on Russian, using the smaller training set on a MEASURE-MONEY rich corpus where the MEASURE and MONEY tokens were previously unseen is shown in Table 13. On the face of it would seem that the FST filter is actually making things worse, until one looks at the differences. Of the 50 cases where the filter made things "worse," 34 (70%) are cases where there was an error in the data and a perfectly well formed measure was rendered with

_______
[11] To remind the reader, all test data are of course held out from the training and development data, but it is common for the same literal expression to recur.

[12] Only in three cases involving Indian Rupees, such as *Rs.149* did the filter match, but still the wrong answer (in this case *six*), was produced. In that case the RNN probably simply failed to produce any paths including the right answer. In such cases the only solution is probably to override the RNN completely on a case-by-case basis.

| Input | RNN | RNN+FST |
|---|---|---|
| £5 | five | five pounds |
| 11 billion AED | eleven billion danish | eleven billion dirhams |
| 2 mA | 2 megaamperes | 2 milliamperes |
| 33 rpm | thirty two revolutions per minute | thirty three revolutions per minute |

Table 11: Some misleading readings of the RNN that have been corrected by the FST.

| | RNN | | RNN+FST filter | |
|---|---|---|---|---|
| ALL | 13152 | 0.983 | 13177 | 0.980 |
| PLAIN | 8176 | 0.998 | 8190 | 0.998 |
| PUNCT | 2501 | 1.000 | 2506 | 1.000 |
| DATE | 130 | 0.969 | 130 | 0.969 |
| TRANS | 165 | 0.970 | 165 | 0.970 |
| LETTERS | 192 | 0.995 | 192 | 0.995 |
| CARDINAL | 435 | 0.931 | 437 | 0.931 |
| VERBATIM | 175 | 1.000 | 175 | 1.000 |
| MEASURE | 1131 | 0.877 | 1133 | 0.856 |
| ORDINAL | 14 | 1.000 | 14 | 1.000 |
| DECIMAL | 49 | 0.939 | 49 | 0.939 |
| DIGIT | 2 | 1.000 | 2 | 1.000 |
| MONEY | 155 | 0.832 | 157 | 0.796 |
| FRACTION | 7 | 0.714 | 7 | 0.714 |
| TIME | 5 | 0.800 | 5 | 0.800 |

Table 13: Accuracies for the attention-based sequence-to-sequence models for Russian on smaller training sets, with and without an FST filter, on the MEASURE-MONEY rich dataset, where in this case all measure and money phrases are previously unseen in the training.

| | RNN | | RNN+FST filter | |
|---|---|---|---|---|
| ALL | 16032 | 0.995 | 16050 | 0.996 |
| PLAIN | 10859 | 0.999 | 10869 | 0.999 |
| PUNCT | 2726 | 1.000 | 2730 | 1.000 |
| DATE | 184 | 1.000 | 184 | 1.000 |
| LETTERS | 167 | 0.964 | 168 | 0.964 |
| CARDINAL | 438 | 0.998 | 439 | 0.998 |
| VERBATIM | 101 | 0.990 | 101 | 0.990 |
| MEASURE | 863 | 0.961 | 865 | 0.979 |
| ORDINAL | 3 | 1.000 | 3 | 1.000 |
| DECIMAL | 196 | 0.995 | 196 | 0.995 |
| ELECTRONIC | 1 | 1.000 | 1 | 1.000 |
| DIGIT | 13 | 1.000 | 13 | 1.000 |
| MONEY | 471 | 0.955 | 471 | 0.955 |
| FRACTION | 7 | 1.000 | 7 | 1.000 |
| TIME | 1 | 1.000 | 1 | 1.000 |
| ADDRESS | 1 | 1.000 | 1 | 1.000 |

Table 12: Accuracies for the attention-based sequence-to-sequence models for English on smaller training sets, with and without an FST filter, on the MEASURE-MONEY rich dataset, where in this case all measure and money phrases are previously unseen in the training. In this case the only improvement of the FST filter was to the MEASURE expressions.

*sil* as the 'truth'. In nearly all other cases, the *input* was actually ill formed and both Kestrel and the RNN without the FST filter 'corrected' the input. For example a Wikipedia contributor wrote *47 292 долларов* '47,292 dollars', which should correctly be *47 292 доллара*, since the preceding number ends in '2', and thus the word for 'dollar' should be in the genitive singular, not the genitive plural. Now, the Kestrel grammars for Russian have the property that they read measure and money expressions, among other semiotic classes, into an internal format that in some cases abstracts away from the written form. In the case at hand the written *долларов* gets represented internally as *dollar*. During the verbalization phase the verbalizer grammars translate this into the form of the word required by the grammatical context, in this case *доллара*. Thus Kestrel has the (one could argue) undesirable property of enforcing grammatical constraints on the input. The result is that the data contains instances of these sorts of corrections where *долларов* gets rendered as *доллара*, and the RNN left to its own devices learns this mapping. Thus the RNN produces *сорок семь тысяч двести девяносто два доллара*. The FST filter, which does not allow *долларов* to be read as *доллара*, verbalizes as written — arguably the right behavior for a TTS system, which should not be in the business of correcting the grammar of the input text. In addition to these cases, there were 67 cases where the RNN+FST was an unequivocal improvement over the RNN alone, as in *10 кН* 'ten kilo-Newtons', which was read by the RNN as *десяти килолитров* 'ten kiloliters' but by the RNN+FST as *десяти килоньютонов* 'ten kilonewtons'. This is of course an instance of a broad class of category errors sometimes made by the RNN alone, that we have seen many instances of.

All in all then, the FST-filtration approach seems to be a viable way to improve the quality of the output for targeted cases where the RNN is prone to make the occasional error.

## 8   Discussion and the challenge

We have presented evidence in this paper that training neural models to learn text normalization is probably not going to be reducible to simply having copious amounts of aligned written- and spoken-form text, and then training a general neural system to compute the mapping. An approach where one combines the RNN with a more knowledge-based system such as an FST, such as we presented in Section 7, is probably a viable approach, but it has yet to be demonstrated that one can do it with RNNs alone.

To be sure, our RNNs were often capable of producing surprisingly good results and learning some complex mappings. Yet they sometimes also produced weird output, making them risky for use in a TTS system. Of course traditional approaches to TTS text normalization make errors, but they are not likely to make an error like reading the wrong number, or substituting *hours* for *gigabytes*, something that the RNNs are quite prone to do. The reason the FST filtering approach works, of course, is precisely because it disallows such random mappings.

Again, the situation is different from some other NLP applications, such as MT or parsing, where deep learning can be used more or less "out-of-the box". Indeed if one were to evaluate a text normalization system on the basis of how well the system does overall, then the systems reported in this paper are already doing very well, with accuracies over 99%. But when one drills down and looks at the interesting cases — say dates, which account for about 2% of the tokens in these data — then the performance is less compelling. An MT system that fails on many instances of a somewhat unusual construction could still be a fairly decent MT system overall. A text normalization system that reads the year *2012* as *two twelve* is seriously problematic no matter how well it does on text overall. Ultimately the difference comes down to different demands of the domain: the bar for text normalization is simply higher.

Given past experience we anticipate three main classes of responses, which we would like to briefly address.

The first is that our characterization of what is important for a text normalization is idiosyncratic: what justification do we have for saying that, for example, a text normalization must get dates correct? But the response to that is obvious: the various *semiotic classes* are precisely where most of the effort has been devoted in developing traditional approaches to text normalization for TTS dating back to the 1970's (Allen et al., 1987), for the simple reason that a TTS system *ought* to be able to know how to read something like *Sep 12, 2014*.

The second is that we have set up a straw man: who ever argued that one could expect a deep learning system to learn a text normalization system from these kind of data? It is true that nobody has specifically made that claim for text normalization, but the view is definitely one that is "in the air": colleagues of one of the authors who work on TTS have been asked why so much hand labor goes into TTS systems. Can one not just get a huge amount of aligned text and speech and learn the mapping? The final and perhaps most anticipated response is: "You didn't use the right kind of models; if you had just used an *X* model with *Y* objective function, etc., then you would have solved the problems you noted." Our response to that is that the data described in this paper will be made publicly available, and people are encouraged to try out their clever ideas for themselves.

The challenge then can be laid out simply as follows: using the data reported here,[13] train a pure deep learning based normalization system for English and Russian that outperforms the results reported in this paper. By "outperform" here we are not primarily focusing on the overall scores, which are already very good, but rather the scores for various of the interesting categories. Rather, can one get a system, for example, that would never read £ as *dollars* or *euros*, or any of the other similar errors where a related but incorrect term has been substituted? (We take it as given that the same training-development-test division of the data is used and the same scoring scripts.) If one could train a pure deep-learning system that failed to make these sorts of silly errors and in general did better than systems reported here on the various semiotic categories, this would represent a true advance over the state of the art reported in this paper.

## Acknowledgements

We thank Alexander Gutkin for preparing the original code for producing Kestrel text normalization, and to Kyle Gorman for producing the data. Alexander Gutkin also checked a sample of the Russian data to estimate Kestrel's error rate. We also thank both Alexander and Kyle, as well as Brian Roark and Suyoun Yoon for comments on earlier versions of this paper. Finally Hasim Sak for help with the LSTM models reported in Experiment 1.

---

[13] Available on `location-to-be-determined`, most likely under the Creative Commons 4.0 license. The data include the training, development and test datasets with their annotated normalized outputs; and scoring scripts. We will also provide a mechanism to submit corrections to the data.